\pdfoutput=1
\documentclass[letterpaper, 10pt, conference]{IEEEtran}
\IEEEoverridecommandlockouts
\usepackage{amsmath,amssymb,amsthm}
\usepackage{graphicx}
\usepackage{booktabs}
\usepackage{xcolor}
\usepackage{tikz}
\usetikzlibrary{arrows.meta,decorations.pathreplacing,calc}
\usepackage{float}
\usepackage{algpseudocode}
\usepackage{cite}
\usepackage[hidelinks]{hyperref}

\floatstyle{ruled}
\newfloat{algorithm}{t}{loa}
\floatname{algorithm}{Algorithm}
\newtheorem{assumption}{Assumption}
\newtheorem{lemma}{Lemma}
\newtheorem{proposition}{Proposition}

\newcommand{\abar}{\bar a}
\newcommand{\SH}{\mathcal{S}_H}
\newcommand{\Sterm}{\mathcal{S}_{\mathrm{term}}}
\newcommand{\vhat}{\hat v}
\newcommand{\chat}{\hat c}

\title{\LARGE \bf Exposure-Window Online Occupancy Calibration and Learned\\ Backups for Safe Navigation Among Moving Obstacles}

\author{Jiashen Hu}

\begin{document}
\maketitle
\thispagestyle{plain}
\pagestyle{plain}

\begin{abstract}
Backup safety filters turn the invariance constraint for safety certification into the safety of finite-horizon forward simulated trajectories that reach and remain inside a terminal safe set. This formulation provides greater flexibility by eliminating the need to construct a global invariant set offline for safety-critical systems. However, the quality of the backup control law or policy strongly affects certification conservatism. We show that this conservatism is governed not only by backup policies, but also by the terminal-set condition over the \emph{exposure window} $[0, t^\star]$---the interval from a candidate action to terminal-set entry---which determines whether task completion is bottlenecked by occupancy bounds alone or by their coupling with backup policy capacity. We formalize two terminal-set conditions for analysis: passive yielding in a hover set $\SH$ versus non-yielding traffic with a static safe zone $\Sterm$. Under these two terminal-set assumptions, we propose a commit-and-verify filter that combines an online per-class feedback tracker calibrating obstacle acceleration bounds over $[0, t^\star]$ with a causal, certificate-aligned learned backup $\pi_b^{\mathrm{NN}}$ trained across the tracker's bound range. We present experiments in Isaac Sim 3D and planar 2D scenarios showing that task performance depends strongly on the terminal-set condition. Ablations show that our learning-based backup also helps compress the \emph{exposure window}. Furthermore, across a $2 \times 2$ benchmark spanning both terminal-set conditions and both planar 2D and Isaac Sim 3D dynamics, our unified filter achieves zero observed collisions and outperforms Gatekeeper-2D/3D while avoiding the collisions of reactive safety shields.
\end{abstract}

\section{Introduction}

Learned navigation policies fly quadrotors through dynamic clutter at low computational cost~\cite{xu2025navrl}, but lack safety guarantees.
A safety filter wraps a nominal policy and overrides its command only when a fallback certificate fails~\cite{hsu2024safetyfilter,wabersich2023datadriven}.
Backup control barrier function (CBF) filters~\cite{gurriet2018online,chen2021backup} and commit-and-verify schemes such as gatekeeper~\cite{agrawal2024gatekeeper} avoid computing a global invariant set offline by certifying a finite rollout into a known terminal safe set.

Among moving obstacles of unknown intent, certification needs to rely on an occupancy model~\cite{fraichard2004ics,kousik2020rtd,kim2026occlusioncbf}, which provides a severe conservatism for filtered action, since the worst-case occupancy inflation over the verification horizon which also causes deadlocks even if there are still feasible manouver for the navigation task.
In our Isaac Sim crowd benchmark, wrapping frozen NavRL~\cite{xu2025navrl} in a worst-case backup filter eliminates all 39 collisions but drops completion from 95.5\,\% to 83.1\,\%.
Prior work mitigates this conservatism mainly along two axes: learning a stronger backup policy to enlarge the certified set~\cite{rabiee2025rlbus,yun2026ps2rl,kiemel2024safe}, or shrinking occupancy via conformal prediction~\cite{lindemann2023safe,dixit2023adaptive,strawn2023cpsf,kim2026ps}. However, in some narrow scenairo, both two kinds of methods lose, which means a stronger backup and a precise occupancy is needed at the same time.
When does shrinking occupancy suffice, and when must occupancy calibration and backup learning act jointly?

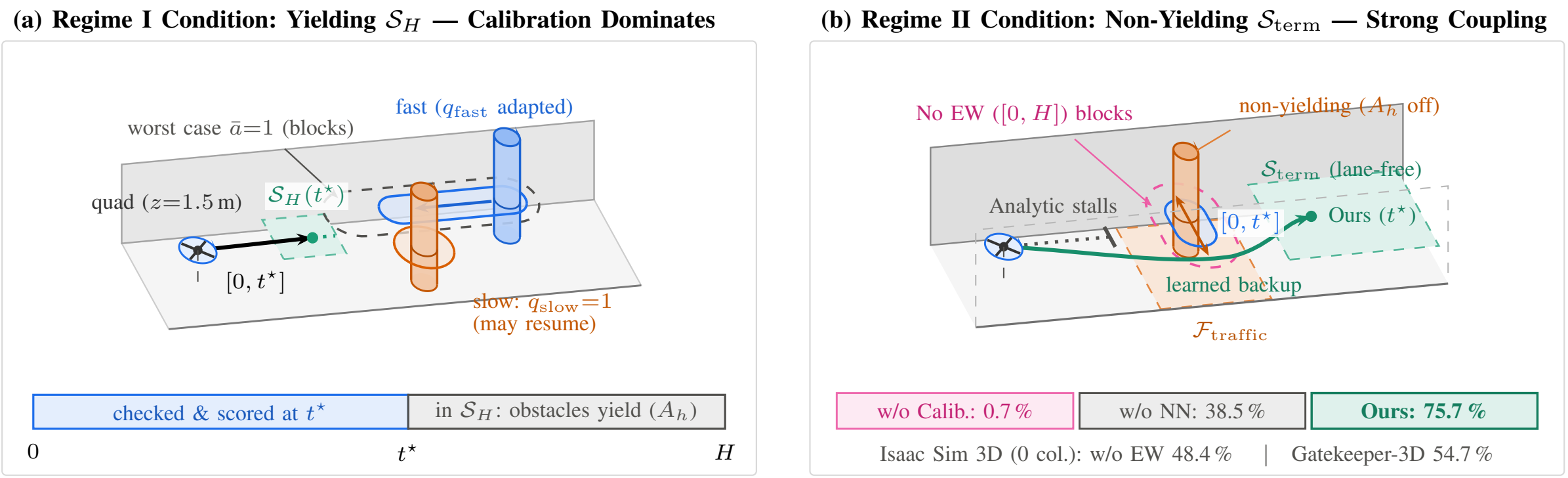
\begin{figure*}[t]
\centering
\begin{tikzpicture}[x=1cm,y=1cm,font=\footnotesize,scale=0.90,transform shape]
  \definecolor{fastc}{HTML}{1F6FEB}
  \definecolor{slowc}{HTML}{D95F02}
  \definecolor{termc}{HTML}{1B9E77}
  \definecolor{lanec}{HTML}{E7298A}
  \definecolor{ink2}{HTML}{52514E}
  \definecolor{wallc}{HTML}{333333}

  \begin{scope}[shift={(0,0)}]
    \node[anchor=west, font=\footnotesize\bfseries] at (0.0, 3.60) {(a) Regime I Condition: Yielding $\mathcal{S}_H$ --- Calibration Dominates};
    \draw[black!15, rounded corners=1.2pt] (0.0, -1.48) rectangle (8.45, 3.38);

    \begin{scope}[shift={(4.25,0.88)}, x={(0.54cm,0.05cm)}, y={(-0.22cm,0.40cm)}, z={(0cm,0.26cm)}]
      \fill[black!4] (-4.9,-1.2,0) -- (4.9,-1.2,0) -- (4.9,1.2,0) -- (-4.9,1.2,0) -- cycle;
      \draw[wallc!25, line width=0.4pt] (-4.9,-1.2,0) -- (4.9,-1.2,0) -- (4.9,1.2,0) -- (-4.9,1.2,0) -- cycle;
      \fill[wallc!12, draw=wallc!45, line width=0.45pt] (-4.9,1.2,0) -- (4.9,1.2,0) -- (4.9,1.2,3.4) -- (-4.9,1.2,3.4) -- cycle;

      \fill[termc!14] (-2.25,-0.55,1.5) -- (-0.95,-0.55,1.5) -- (-0.95,0.55,1.5) -- (-2.25,0.55,1.5) -- cycle;
      \draw[termc!70, dashed, line width=0.5pt] (-2.25,-0.55,1.5) -- (-0.95,-0.55,1.5) -- (-0.95,0.55,1.5) -- (-2.25,0.55,1.5) -- cycle;

      \draw[ink2, dashed, line width=0.65pt]
        plot[domain=-90:90, samples=18] ({2.80 + 0.62*cos(\x)}, {0.48 + 0.62*sin(\x)}, 1.5) --
        plot[domain=90:270, samples=18] ({-0.35 + 0.62*cos(\x)}, {0.48 + 0.62*sin(\x)}, 1.5) -- cycle;

      \fill[fastc!16, fill opacity=0.72]
        plot[domain=-90:90, samples=18] ({2.80 + 0.38*cos(\x)}, {0.48 + 0.38*sin(\x)}, 1.5) --
        plot[domain=90:270, samples=18] ({0.55 + 0.38*cos(\x)}, {0.48 + 0.38*sin(\x)}, 1.5) -- cycle;
      \draw[fastc, line width=0.7pt]
        plot[domain=-22:90, samples=14] ({2.80 + 0.38*cos(\x)}, {0.48 + 0.38*sin(\x)}, 1.5) --
        plot[domain=90:158, samples=10] ({0.55 + 0.38*cos(\x)}, {0.48 + 0.38*sin(\x)}, 1.5);

      \fill[fastc!26] plot[domain=0:360, samples=32, smooth cycle] ({2.80 + 0.25*cos(\x)}, {0.48 + 0.25*sin(\x)}, 0);
      \fill[fastc!34, opacity=0.90]
        plot[domain=158:338, samples=18] ({2.80 + 0.25*cos(\x)}, {0.48 + 0.25*sin(\x)}, 0) --
        ({2.80 + 0.25*cos(338)}, {0.48 + 0.25*sin(338)}, 4.2) --
        plot[domain=338:158, samples=18] ({2.80 + 0.25*cos(\x)}, {0.48 + 0.25*sin(\x)}, 4.2) -- cycle;
      \draw[fastc!90!black, line width=0.65pt]
        ({2.80 + 0.25*cos(158)}, {0.48 + 0.25*sin(158)}, 0) -- ({2.80 + 0.25*cos(158)}, {0.48 + 0.25*sin(158)}, 4.2)
        ({2.80 + 0.25*cos(338)}, {0.48 + 0.25*sin(338)}, 0) -- ({2.80 + 0.25*cos(338)}, {0.48 + 0.25*sin(338)}, 4.2);
      \draw[fastc!90!black, line width=0.55pt] plot[domain=158:338, samples=18] ({2.80 + 0.25*cos(\x)}, {0.48 + 0.25*sin(\x)}, 0);
      \draw[fastc!80!black, line width=0.5pt] plot[domain=158:338, samples=18] ({2.80 + 0.25*cos(\x)}, {0.48 + 0.25*sin(\x)}, 1.5);
      \fill[fastc!48, draw=fastc!90!black, line width=0.6pt]
        plot[domain=0:360, samples=32, smooth cycle] ({2.80 + 0.25*cos(\x)}, {0.48 + 0.25*sin(\x)}, 4.2);
      \draw[fastc, line width=0.7pt]
        plot[domain=158:270, samples=14] ({0.55 + 0.38*cos(\x)}, {0.48 + 0.38*sin(\x)}, 1.5) --
        plot[domain=-90:-22, samples=10] ({2.80 + 0.38*cos(\x)}, {0.48 + 0.38*sin(\x)}, 1.5);
      \draw[-{Stealth[length=4pt]}, fastc!90!black, line width=0.85pt] (2.48,0.48,1.5) -- (0.88,0.48,1.5);

      \fill[slowc!15, fill opacity=0.72]
        plot[domain=0:360, samples=36, smooth cycle] ({0.65 + 0.58*cos(\x)}, {-0.58 + 0.58*sin(\x)}, 1.5);
      \draw[slowc, line width=0.65pt]
        plot[domain=-22:158, samples=22] ({0.65 + 0.58*cos(\x)}, {-0.58 + 0.58*sin(\x)}, 1.5);
      \fill[slowc!26] plot[domain=0:360, samples=32, smooth cycle] ({0.65 + 0.25*cos(\x)}, {-0.58 + 0.25*sin(\x)}, 0);
      \fill[slowc!35, opacity=0.90]
        plot[domain=158:338, samples=18] ({0.65 + 0.25*cos(\x)}, {-0.58 + 0.25*sin(\x)}, 0) --
        ({0.65 + 0.25*cos(338)}, {-0.58 + 0.25*sin(338)}, 3.9) --
        plot[domain=338:158, samples=18] ({0.65 + 0.25*cos(\x)}, {-0.58 + 0.25*sin(\x)}, 3.9) -- cycle;
      \draw[slowc!90!black, line width=0.65pt]
        ({0.65 + 0.25*cos(158)}, {-0.58 + 0.25*sin(158)}, 0) -- ({0.65 + 0.25*cos(158)}, {-0.58 + 0.25*sin(158)}, 3.9)
        ({0.65 + 0.25*cos(338)}, {-0.58 + 0.25*sin(338)}, 0) -- ({0.65 + 0.25*cos(338)}, {-0.58 + 0.25*sin(338)}, 3.9);
      \draw[slowc!90!black, line width=0.55pt] plot[domain=158:338, samples=18] ({0.65 + 0.25*cos(\x)}, {-0.58 + 0.25*sin(\x)}, 0);
      \draw[slowc!80!black, line width=0.5pt] plot[domain=158:338, samples=18] ({0.65 + 0.25*cos(\x)}, {-0.58 + 0.25*sin(\x)}, 1.5);
      \fill[slowc!50, draw=slowc!90!black, line width=0.6pt]
        plot[domain=0:360, samples=32, smooth cycle] ({0.65 + 0.25*cos(\x)}, {-0.58 + 0.25*sin(\x)}, 3.9);
      \draw[slowc, line width=0.70pt]
        plot[domain=158:338, samples=22] ({0.65 + 0.58*cos(\x)}, {-0.58 + 0.58*sin(\x)}, 1.5);

      \coordinate (q1drone) at (-3.85, -0.10, 1.5);
      \coordinate (q1stop)  at (-1.45, -0.05, 1.5);
      \draw[line width=1.15pt, -{Stealth[length=4.5pt]}] (q1drone) .. controls (-2.85, -0.08, 1.5) and (-2.05, -0.06, 1.5) .. (q1stop);
      \draw[termc, dotted, line width=1.0pt] (q1stop) -- (-0.98, -0.05, 1.5);
      \fill[termc] (q1stop) circle (1.8pt);

      \fill[fastc!12, draw=fastc, line width=0.55pt]
        plot[domain=0:360, samples=32, smooth cycle] ({-3.85 + 0.36*cos(\x)}, {-0.10 + 0.36*sin(\x)}, 1.5);
      \draw[wallc, line width=0.75pt] (-4.10,-0.32,1.5) -- (-3.60,0.12,1.5) (-4.10,0.12,1.5) -- (-3.60,-0.32,1.5);
      \fill[wallc] (q1drone) circle (1.5pt);
      \draw[ink2, dashed, line width=0.4pt] (q1drone) -- (-3.85, -0.10, 0);
      \draw[wallc!65, line width=0.55pt] (-4.9,-1.2,0) -- (4.9,-1.2,0);

      \node[anchor=south, font=\scriptsize, ink2] at (-2.45, 1.15, 3.65) {worst case $\bar a{=}1$ (blocks)};
      \draw[ink2, -{Stealth[length=3pt]}, line width=0.45pt] (-1.55, 1.12, 3.55) -- (-0.55, 0.85, 1.65);
      \node[anchor=south, font=\scriptsize, fastc!90!black] at (2.40, 0.65, 4.35) {fast ($q_{\mathrm{fast}}$ adapted)};
      \node[anchor=north west, font=\scriptsize, slowc!90!black, align=left] at (1.28, -0.95, 0.45) {slow: $q_{\mathrm{slow}}{=}1$\\[-1pt](may resume)};
      \node[anchor=south, font=\scriptsize, wallc] at (-4.35, 0.25, 2.15) {quad ($z{=}1.5$\,m)};
      \node[anchor=south, font=\scriptsize, termc!85!black, fill=white, fill opacity=0.85, text opacity=1, inner sep=0.8pt] at (-1.35, 0.48, 1.85) {$\mathcal{S}_H(t^\star)$};
      \node[anchor=north, font=\scriptsize] at (-2.75, -0.35, 1.25) {$[0,t^\star]$};
    \end{scope}

    \fill[fastc!12] (0.35, -0.96) rectangle (4.55, -0.58);
    \fill[black!7] (4.55, -0.96) rectangle (8.10, -0.58);
    \draw[fastc, line width=0.6pt] (0.35, -0.96) rectangle (4.55, -0.58);
    \draw[ink2, line width=0.6pt] (4.55, -0.96) rectangle (8.10, -0.58);
    \node[fastc!80!black, font=\scriptsize] at (2.45, -0.77) {checked \& scored at $t^\star$};
    \node[ink2, font=\scriptsize] at (6.32, -0.77) {in $\mathcal{S}_H$: obstacles yield ($A_h$)};
    \node[anchor=north, font=\scriptsize] at (0.35, -1.01) {$0$};
    \node[anchor=north, font=\scriptsize] at (4.55, -1.01) {$t^\star$};
    \node[anchor=north, font=\scriptsize] at (8.10, -1.01) {$H$};
  \end{scope}

  \begin{scope}[shift={(9.05,0)}]
    \node[anchor=west, font=\footnotesize\bfseries] at (0.0, 3.60) {(b) Regime II Condition: Non-Yielding $\mathcal{S}_{\mathrm{term}}$ --- Strong Coupling};
    \draw[black!15, rounded corners=1.2pt] (0.0, -1.48) rectangle (8.45, 3.38);

    \begin{scope}[shift={(4.25,0.88)}, x={(0.54cm,0.05cm)}, y={(-0.22cm,0.40cm)}, z={(0cm,0.26cm)}]
      \fill[black!4] (-4.9,-1.15,0) -- (4.9,-1.15,0) -- (4.9,1.15,0) -- (-4.9,1.15,0) -- cycle;
      \fill[wallc!16, draw=wallc!60, line width=0.45pt] (-4.9,1.15,0) -- (4.9,1.15,0) -- (4.9,1.15,4.2) -- (-4.9,1.15,4.2) -- cycle;
      \fill[slowc!15] (-1.05,-1.12,0) -- (1.25,-1.12,0) -- (1.25,1.12,0) -- (-1.05,1.12,0) -- cycle;
      \draw[slowc!75, dashed, line width=0.5pt] (-1.05,-1.12,0) -- (1.25,-1.12,0) -- (1.25,1.12,0) -- (-1.05,1.12,0) -- cycle;
      \fill[termc!14] (1.65,-1.05,1.5) -- (4.75,-1.05,1.5) -- (4.75,1.05,1.5) -- (1.65,1.05,1.5) -- cycle;
      \draw[termc!75, dashed, line width=0.5pt] (1.65,-1.05,1.5) -- (4.75,-1.05,1.5) -- (4.75,1.05,1.5) -- (1.65,1.05,1.5) -- cycle;

      \draw[lanec!80, dashed, line width=0.68pt]
        plot[domain=0:180, samples=18] ({0.10 + 0.72*cos(\x)}, {0.55 + 0.72*sin(\x)}, 1.5) --
        plot[domain=180:360, samples=18] ({0.10 + 0.72*cos(\x)}, {-0.35 + 0.72*sin(\x)}, 1.5) -- cycle;

      \fill[fastc!16, fill opacity=0.74]
        plot[domain=0:180, samples=18] ({0.10 + 0.40*cos(\x)}, {0.42 + 0.40*sin(\x)}, 1.5) --
        plot[domain=180:360, samples=18] ({0.10 + 0.40*cos(\x)}, {-0.08 + 0.40*sin(\x)}, 1.5) -- cycle;
      \draw[fastc, line width=0.68pt]
        plot[domain=0:180, samples=20] ({0.10 + 0.40*cos(\x)}, {0.42 + 0.40*sin(\x)}, 1.5);

      \fill[slowc!26] plot[domain=0:360, samples=32, smooth cycle] ({0.10 + 0.24*cos(\x)}, {0.42 + 0.24*sin(\x)}, 0);
      \fill[slowc!35, opacity=0.90]
        plot[domain=158:338, samples=18] ({0.10 + 0.24*cos(\x)}, {0.42 + 0.24*sin(\x)}, 0) --
        ({0.10 + 0.24*cos(338)}, {0.42 + 0.24*sin(338)}, 4.2) --
        plot[domain=338:158, samples=18] ({0.10 + 0.24*cos(\x)}, {0.42 + 0.24*sin(\x)}, 4.2) -- cycle;
      \draw[slowc!90!black, line width=0.65pt]
        ({0.10 + 0.24*cos(158)}, {0.42 + 0.24*sin(158)}, 0) -- ({0.10 + 0.24*cos(158)}, {0.42 + 0.24*sin(158)}, 4.2)
        ({0.10 + 0.24*cos(338)}, {0.42 + 0.24*sin(338)}, 0) -- ({0.10 + 0.24*cos(338)}, {0.42 + 0.24*sin(338)}, 4.2);
      \draw[slowc!90!black, line width=0.55pt] plot[domain=158:338, samples=18] ({0.10 + 0.24*cos(\x)}, {0.42 + 0.24*sin(\x)}, 0);
      \draw[slowc!80!black, line width=0.5pt] plot[domain=158:338, samples=18] ({0.10 + 0.24*cos(\x)}, {0.42 + 0.24*sin(\x)}, 1.5);
      \fill[slowc!50, draw=slowc!90!black, line width=0.6pt]
        plot[domain=0:360, samples=32, smooth cycle] ({0.10 + 0.24*cos(\x)}, {0.42 + 0.24*sin(\x)}, 4.2);
      \draw[fastc, line width=0.68pt]
        ({0.10 - 0.40}, 0.42, 1.5) --
        plot[domain=180:360, samples=18] ({0.10 + 0.40*cos(\x)}, {-0.08 + 0.40*sin(\x)}, 1.5) --
        ({0.10 + 0.40}, 0.42, 1.5);
      \draw[{Stealth[length=3.5pt]}-{Stealth[length=3.5pt]}, slowc!90!black, line width=0.75pt] (0.10,-0.75,1.5) -- (0.10,0.95,1.5);

      \coordinate (q2drone) at (-3.85, 0.0, 1.5);
      \coordinate (q2term)  at (2.55, 0.05, 1.5);
      \draw[ink2, dotted, line width=0.95pt] (q2drone) -- (-1.65, 0.0, 1.5);
      \draw[ink2, line width=1.1pt] (-1.65, -0.28, 1.5) -- (-1.65, 0.28, 1.5);
      \draw[termc!90!black, line width=1.15pt, -{Stealth[length=4.5pt]}]
        (q2drone) .. controls (-2.10, -0.35, 1.5) and (-0.75, -0.92, 1.5) .. (0.55, -0.82, 1.5) .. controls (1.35, -0.68, 1.5) and (1.90, -0.10, 1.5) .. (q2term);
      \fill[termc!90!black] (q2term) circle (1.8pt);

      \fill[fastc!12, draw=fastc, line width=0.55pt]
        plot[domain=0:360, samples=32, smooth cycle] ({-3.85 + 0.36*cos(\x)}, {0.0 + 0.36*sin(\x)}, 1.5);
      \draw[wallc, line width=0.75pt] (-4.10,-0.22,1.5) -- (-3.60,0.22,1.5) (-4.10,0.22,1.5) -- (-3.60,-0.22,1.5);
      \fill[wallc] (q2drone) circle (1.5pt);
      \draw[ink2, dashed, line width=0.4pt] (q2drone) -- (-3.85, 0.0, 0);
      \draw[wallc!70, line width=0.6pt] (-4.9,-1.15,0) -- (4.9,-1.15,0);
      \draw[wallc!35, dashed, line width=0.4pt] (-4.9,-1.15,0) -- (-4.9,-1.15,4.2) -- (4.9,-1.15,4.2) -- (4.9,-1.15,0);

      \node[anchor=south, font=\scriptsize, lanec!85!black] at (-2.95, 1.15, 4.30) {No EW ($[0,H]$) blocks};
      \draw[lanec!75, -{Stealth[length=3pt]}, line width=0.45pt] (-1.95, 1.05, 4.15) -- (-0.52, 0.62, 1.65);
      \node[anchor=south west, font=\scriptsize, slowc!90!black] at (1.15, 0.85, 4.35) {non-yielding ($A_h$ off)};
      \draw[slowc!80, line width=0.45pt] (1.18, 0.80, 4.32) -- (0.32, 0.44, 4.15);
      \node[anchor=south, font=\scriptsize, termc!80!black, align=center] at (3.35, 0.48, 1.78) {$\mathcal{S}_{\mathrm{term}}$ (lane-free)};
      \node[anchor=north, font=\scriptsize, slowc!85!black] at (0.35, -1.25, -0.15) {$\mathcal{F}_{\mathrm{traffic}}$};
      \node[anchor=south, font=\scriptsize, ink2] at (-2.75, 0.18, 1.78) {Analytic stalls};
      \node[anchor=north, font=\scriptsize, termc!80!black] at (0.55, -0.92, 1.25) {learned backup};
      \node[anchor=west, font=\scriptsize, termc!80!black] at (2.68, 0.05, 1.50) {Ours ($t^\star$)};
      \node[anchor=west, font=\scriptsize, fastc, fill=white, fill opacity=0.85, text opacity=1, inner sep=0.8pt] at (0.62, 0.12, 1.48) {$[0,t^\star]$};
    \end{scope}

    \fill[lanec!10] (0.30, -0.96) rectangle (2.95, -0.56);
    \fill[black!7] (3.02, -0.96) rectangle (5.55, -0.56);
    \fill[termc!14] (5.62, -0.96) rectangle (8.15, -0.56);
    \draw[lanec!70, line width=0.55pt] (0.30, -0.96) rectangle (2.95, -0.56);
    \draw[ink2, line width=0.55pt] (3.02, -0.96) rectangle (5.55, -0.56);
    \draw[termc!80!black, line width=0.65pt] (5.62, -0.96) rectangle (8.15, -0.56);
    \node[lanec!85!black, font=\scriptsize] at (1.625, -0.76) {w/o Calib.: 0.7\,\%};
    \node[ink2, font=\scriptsize] at (4.285, -0.76) {w/o NN: 38.5\,\%};
    \node[termc!80!black, font=\scriptsize\bfseries] at (6.885, -0.76) {Ours: 75.7\,\%};
    \node[anchor=north, font=\scriptsize, ink2] at (4.22, -1.01) {Isaac Sim 3D (0 col.): w/o EW 48.4\,\% \quad\textbar\quad Gatekeeper-3D 54.7\,\%};
  \end{scope}
\end{tikzpicture}
\vspace{-4pt}
\caption{Exposure window $[0, t^\star]$ under the two terminal-set conditions. Panel~(a): under passive yielding in $\SH$ (Regime~I), stopping in place is safe; online calibration shrinks $q_{\mathrm{fast}}$ while holding $q_{\mathrm{slow}}{=}1$, decoupling calibration from backup choice. Panel~(b): under non-yielding traffic with $\Sterm$ excluding $\mathcal{F}_{\mathrm{traffic}}$ (Regime~II), exposure-window calibration and $\pi_b^{\mathrm{NN}}$ couple strongly to reach $\Sterm$.}\label{fig:concept}
\vspace{-6pt}
\end{figure*}

We show that the answer is governed by the \emph{exposure window} $[0, t^\star]$---the time from a candidate action until terminal-set entry---as illustrated in Fig.~\ref{fig:concept}.
Must to be reminded, the terminal set regime is not the source of conservatism; it acts as the structural condition that determines which mechanism dominates conservatism and how calibration interacts with backup capacity.
Under \emph{Regime~I} (passive yielding in a hover set $\SH$), braking to a stop is always terminal-safe and locks $t^\star \approx 1.2\text{--}1.3$\,s across controllers; lateral escapes cannot clear the worst-case occupancy bubble within $t^\star$, so analytic and learned backups perform alike while online calibration of $\abar$ over $[0, t^\star]$ recovers most of the lost completion.
Under \emph{Regime~II} (non-yielding traffic), hovering in traffic is unsafe, so forward invariance requires a static safe zone $\Sterm$ outside swept traffic lanes.
Under $\Sterm$, occupancy calibration and backup learning become strongly coupled: truncating verification from $[0, H]$ to $[0, t^\star]$, adapting per-class bounds online, and pairing them with a certificate-aligned learned backup $\pi_b^{\mathrm{NN}}$ are all required to reach $\Sterm$.

Calibrating $\abar$ over $[0, t^\star]$ requires four design choices:
first, decision-conditional scoring on committed plans over $[0, t^\star]$;
second, online, on-policy tracking to absorb performative shifts~\cite{perdomo2020performative} such as yield-release restarts;
third, per-speed-class bounds to separate slow and fast risk asymmetry; and fourth, causal early finalization at $t^\star$ so calibration feedback matches the plan's true exposure horizon across control rates.

\textbf{Our contributions:}
\begin{enumerate}
\item \emph{Terminal-set regime taxonomy and exposure-window analysis:} We categorize backup-filter terminal assumptions into passive yielding ($\SH$, Regime~I) versus traffic-excluding safe zones ($\Sterm$, Regime~II) and formalize the exposure window $[0, t^\star]$ in Lemma~\ref{lem:exposure}, explaining how the terminal-set condition shifts the dominant conservatism bottleneck and component coupling.
\item \emph{Online exposure window and learned backup:} We combine a decision-conditional per-class online tracker scored and finalized at $t^\star$ with a causal learned backup $\pi_b^{\mathrm{NN}}$ parameterized and trained directly against the online exposure-window certificate~\eqref{eq:backup_param}--\eqref{eq:backup_loss}.
\item \emph{Systematic and cross-fidelity validation:} In Isaac Sim simulations, controlled $2{\times}2$ ablations show that online calibration improves analytic and learned backups alike under $\SH$, whereas calibration and $\pi_b^{\mathrm{NN}}$ must work together to achieve higher completion under $\Sterm$; across a $2{\times}2$ benchmark spanning both conditions and both 2D and 3D dynamics, our filter maintains zero collisions while outperforming Gatekeeper-2D/3D and avoiding the collisions of reactive shields.
\end{enumerate}

\section{Related Work}

\textbf{Backup CBFs and commit-and-verify filters.}
Standard CBF-QPs~\cite{ames2017cbf,ames2019cbf} require a valid control-invariant barrier offline and can become infeasible or myopic under actuator limits in cluttered dynamic scenes.
Backup CBFs~\cite{gurriet2018online,gurriet2020scalable,chen2021backup}, gatekeeper~\cite{agrawal2024gatekeeper}, and OcclusionCBF~\cite{kim2026occlusioncbf} avoid computing global invariant sets by certifying finite rollouts into a terminal safe set, yet rely on fixed analytic backups and conservative obstacle bounds over the full horizon $[0, H]$, which triggers deadlocks in narrow non-yielding corridors.
PSS~\cite{kim2026ps} pairs split-conformal occupancy with backup rollouts, but calibrates offline over fixed time grids and thus cannot adapt on-policy to motion shifts or match variable entry times.
We instead score committed plans directly over their variable exposure window $[0, t^\star]$ and adapt a physical acceleration bound online per speed class.

\textbf{Learning the backup policy.}
Shielding~\cite{alshiekh2018shield} and CBF layers~\cite{cheng2019endtoend} protect RL policies but rely on discrete abstractions or analytical barriers that scale poorly to uncertain moving obstacles.
RLBUS~\cite{rabiee2025rlbus}, PS2-RL~\cite{yun2026ps2rl}, and Kiemel et al.~\cite{kiemel2024safe} instead train backup policies to enlarge certified sets, but treat obstacle uncertainty as fixed and do not identify when backup learning actually helps.
We show that this depends on the terminal-set condition: learned backups bring negligible gain under yielding hover sets $\SH$ where worst-case occupancy dominates, yet become essential under non-yielding $\Sterm$ once coupled with exposure-window calibration.

\textbf{Motion safety and conformal control.}
Worst-case methods such as inevitable collision states~\cite{fraichard2004ics} and reachability-based planning~\cite{kousik2020rtd} prevent collisions but suffer from freezing-robot deadlocks as worst-case reachable sets grow rapidly with horizon.
Passive and not-at-fault safety~\cite{bouraine2012passive,mitsch2017verification,vaskov2019notatfault} ease conservatism by assuming obstacles yield to a stopped robot ($\SH$, Regime~I), an assumption that fails in non-yielding traffic ($\Sterm$, Regime~II).
Conformal prediction~\cite{vovk2005alrw,vovk2012conditional} and adaptive variants including ACI~\cite{gibbs2021aci}, conformal PID~\cite{angelopoulos2023pid}, and multivalid calibration~\cite{bastani2022multivalid} tighten uncertainty sets in MPC~\cite{lindemann2023safe}, planning~\cite{dixit2023adaptive}, and RL filters~\cite{strawn2023cpsf}; however, offline calibration degrades under closed-loop performative shift~\cite{perdomo2020performative}, while fixed-horizon or class-pooled bounds either over-inflate past terminal-set entry or under-cover abrupt slow-obstacle restarts.
On-policy schemes such as CDT~\cite{lekeufack2024cdt}, decentralized conformal CBFs~\cite{huriot2025multiagent}, and ACoFi~\cite{huriot2026acofi} update abstract decision or CBF thresholds under shift, whereas we directly adapt the physical acceleration bound of a backup capsule over variable $[0, t^\star]$ with entry-step finalization.

\section{Certified Backup Filter and Exposure Window}\label{sec:setting}

\textbf{System and terminal-set conditions.}
A nominal policy outputs $u^{\mathrm{nom}}_k$ at control step $\Delta$ over an $N$-step verification horizon $H{=}N\Delta$, rolled out under a prediction model with bounded state-mismatch error tubes $(e_p, e_v, e_z)$.
A commit-and-verify filter~\cite{agrawal2024gatekeeper} certifies nominal-prefix plus backup $\pi_b$ rollouts into a terminal set and retains the stored plan $\mathcal{P}_{\mathrm{buf}}$ on failure via Algorithm~\ref{alg:unified}; any step updating $\mathcal{P}_{\mathrm{buf}}$ is a \emph{commit step}.
\begin{figure*}[t]
\centering
\includegraphics[width=0.96\textwidth]{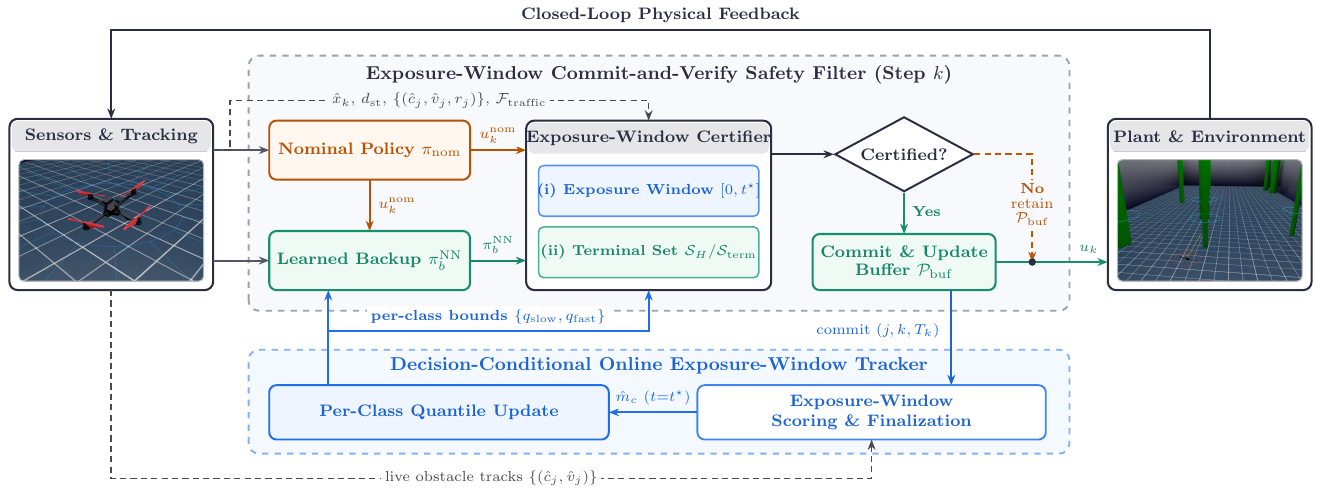}
\vspace{-4pt}
\caption{Closed-loop control architecture combining the exposure-window commit-and-verify safety filter, certificate-aligned learned backup $\pi_b^{\mathrm{NN}}$, and decision-conditional online per-class occupancy tracker.}\label{fig:control_flow}
\vspace{-4pt}
\end{figure*}

To examine how the dominant mechanism of filter conservatism shifts between worst-case occupancy bounds and backup escape capability, we formalize two terminal-set conditions:
\begin{itemize}
\item \emph{Regime~I, yielding hover set $\SH$:} States satisfy $\|\hat v\| + e_v(t) \le v_{\mathrm{stop}}$ and static clearance $d_{\mathrm{st}}(\hat p) \ge r_d + \delta_W + e_p(t)$ with robot radius $r_d$ and hover margin $\delta_W$, under passive yielding:
\begin{assumption}[Yielding, $A_h$]\label{as:yield}
Dynamic obstacles do not collide with a robot stopped inside $\SH$.
\end{assumption}
\item \emph{Regime~II, non-yielding safe set $\Sterm$:} Dropping $A_h$, obstacles never yield. Because global forward invariance cannot be maintained in general against encirclement without a traffic-free refuge, Regime~II assumes a static safe zone outside the traffic envelope $\mathcal{F}_{\mathrm{traffic}}$:
\begin{assumption}[Static safe zone, $A_{\mathrm{safe}}$]\label{as:safezone}
The workspace admits a non-empty static safe set $\Sterm = \SH \cap \{x : d_{\mathcal{F}_{\mathrm{traffic}}}(p) \ge r_d + d_{\mathrm{brake}}(v)\}$ outside $\mathcal{F}_{\mathrm{traffic}}$, within which a stopped vehicle remains permanently collision-free.
\end{assumption}
Here $\mathcal{F}_{\mathrm{traffic}}$ is a purely spatial path bounding envelope without future obstacle timestamps, supplied identically to all methods.
\end{itemize}

\textbf{Occupancy capsule and exposure-window certificate.}
Sensors report noisy states with persistent track IDs for up to $K_{\mathrm{ret}}$ nearest obstacles within $R_s$, with design noise bounds $\varepsilon$ and $\varepsilon_v$.
For obstacle $j$ of radius $r_j$, sensed centre $\chat_j$, and velocity $\vhat_j$, predicted centre occupancy at lag $t>0$ is the intersection of a speed disc with speed bound $\bar v$ and an acceleration-bounded capsule:
\begin{equation}
\mathcal{O}_j(t;\abar) = \mathcal{B}(\chat_j,\varepsilon{+}\bar v t) \cap \big(\{\chat_j{+}s\vhat_j : s{\in}[0,t]\} \oplus \mathcal{B}(\rho(t;\abar))\big),
\label{eq:capsule}
\end{equation}
where $\rho(t;\abar) = \varepsilon + \varepsilon_v t + \tfrac12 \abar t^2$ for an obstacle acceleration bound $\abar$.
Let $\delta_\Delta = (v_{R,\max} + \bar v)\Delta/2$ bound inter-sample sweep, and let $R_{\mathrm{eff}}(k) = \min(R_s, \|\chat_{(K_{\mathrm{ret}})}-p_k\|)$ when all slots are full and $R_s$ otherwise.
A plan $\{\hat p_i, \hat v_i\}_{i=0}^N$ certifies with entry index $i^\star \in [i_{\min}, N]$ if:
(i)~for all $0 \le i \le i^\star$, it maintains static clearance $d_{\mathrm{st}}(\hat p_i) \ge r_d + e_p(i\Delta) + \delta_\Delta$, altitude bounds, dynamic clearance from $\mathcal{O}_j(i\Delta;\abar_j) \oplus \mathcal{B}(r_j + r_d + e_p(i\Delta) + \delta_\Delta)$, and the unmodeled-obstacle range bound
\begin{equation}
\|\hat p_i - p_k\| \le R_{\mathrm{eff}}(k) - \varepsilon - r_{\max} - r_d - \bar v\, i\Delta - e_p(i\Delta) - \delta_\Delta;
\label{eq:n1}
\end{equation}
and (ii)~for all $i^\star \le i \le N$, it remains inside the terminal set $\SH$ or $\Sterm$. We call $[0, t^\star]$ with $t^\star = i^\star\Delta$ the \emph{exposure window}.

\begin{lemma}[Exposure window]\label{lem:exposure}
Suppose a committed plan certifies with entry $t^\star$ under bounds $\abar_j$, true states obey $(e_p, e_v, e_z)$, sensing errors obey $(\varepsilon, \varepsilon_v)$, obstacles obey $(\bar v, r_{\max})$, inter-sample loss obeys $\delta_\Delta$, and Assumption~\ref{as:yield} or~\ref{as:safezone} holds. If every retained obstacle $j$ satisfies $p_j(t_k+t) \in \mathcal{O}_j(t;\abar_j)$ for all $t \in [0, t^\star]$, the executed trajectory on $[0, H]$ is free of dynamic collisions.
\end{lemma}
\begin{proof}[Proof sketch]
Dynamic clearance and~\eqref{eq:n1} with $\delta_\Delta$ guarantee collision freedom on $[0, t^\star]$ for retained and omitted obstacles, while condition~(ii) keeps $t \in [t^\star, H]$ in the terminal set, where Assumption~\ref{as:yield} or~\ref{as:safezone} prevents contact.
\end{proof}

\textbf{How the terminal-set condition shapes conservatism and coupling.}
By Lemma~\ref{lem:exposure}, $\abar_j$ acts \emph{only} on $[0, t^\star]$ through the quadratic term $\tfrac12\abar t^2$.
Under yielding in $\SH$, forward braking stops anywhere clear of walls within a brief exposure window $t^\star$; sideways escape displaces too little to clear the worst-case capsule $\rho(t^\star; 1)$, so backup choice has negligible impact while calibrating $\abar$ enters quadratically and restores feasible transitions.
Once yielding is removed under $\Sterm$, braking stalls outside $\mathcal{F}_{\mathrm{traffic}}$ and full-horizon certification over $[0, H]$ blocks corridor passage; safe traversal therefore couples exposure-window truncation, online calibration, and a learned backup $\pi_b^{\mathrm{NN}}$ together.

\section{Method: Exposure-Window Calibration and Learned Backups}\label{sec:method}

Fig.~\ref{fig:control_flow} summarizes the closed-loop architecture coupling the exposure-window safety filter, learned backup $\pi_b^{\mathrm{NN}}$, and online per-class occupancy tracker.

\subsection{Decision-Conditional Online Exposure-Window Tracker}
For retained obstacle $j$ at commit $k$, its ground-truth capsule score over $\mathcal{T}_k = \{\Delta, \ldots, t^\star\}$ is $s_{jk} = \max_{t \in \mathcal{T}_k} 2(\delta_{jk}(t) - \varepsilon - \varepsilon_v t)_+ / t^2$, where $\delta_{jk}(t)$ is distance to the capsule spine $\{\chat_j + s\vhat_j : s\in[0,t]\}$; a committed plan incurs a ground-truth \emph{Plan Miss} if any retained obstacle satisfies $s_{jk} > \abar_j$.
Online, Algorithm~\ref{alg:unified} maintains one bound $q_c \in [0, q_{\max}]$ with $q_{\max}{=}1.0$ per speed class---slow $c{=}0$ for $\|\vhat_j\|<0.5$\,m/s and fast $c{=}1$ otherwise---and computes the sensed acceleration nonconformity score over $\mathcal{T}_k$:
\begin{equation}
\hat s_{jk} = \max_{t \in \mathcal{T}_k} \frac{2\,(\hat\delta_{jk}(t) - 2\varepsilon - \varepsilon_v t)_+}{t^2}.
\label{eq:sensed}
\end{equation}
Because Lemma~\ref{lem:exposure} terminates dynamic dependence at $t^\star$, each open record committed at $k'$ finalizes immediately at age $k - k' = k^\star_{k'}$, halving feedback delay without waiting for $H$.
Unmatched tracks exiting $R_s$ receive no further updates and close as covered.
At every step where class-$c$ records finalize with online \emph{Tracker Miss} fraction $\hat m_c$ (records with $\hat s_{jk'} > q_{c(j)}$), $q_c$ updates via quantile tracking~\cite{gibbs2021aci,angelopoulos2023pid}:
\begin{equation}
q_c \leftarrow \operatorname{clip}_{[0, q_{\max}]}\big(q_c + \eta\,(\hat m_c - \alpha)\big).
\label{eq:update}
\end{equation}

\begin{algorithm}[t]
\caption{Exposure-window filter \& Online tracker}\label{alg:unified}
\begin{algorithmic}[1]
\Require $u^{\mathrm{nom}}_k$, stored plan $\mathcal{P}_{\mathrm{buf}}$, class bounds $\{q_c\}$
\State Set $\abar_j \gets q_{c(\|\vhat_j\|)}$ for sensed obstacles $j$ \Comment{capsule bounds}
\State Roll out nominal prefixes $+\pi_b$; certify via~(i)--(ii) under $\{\abar_j\}$
\If{nominal rollout certifies with entry $i^\star$}
  \State $\mathcal{P}_{\mathrm{buf}} \gets \mathcal{P}^{\mathrm{nom}}$; $\mathrm{commit} \gets \mathrm{true}$; $T_k \gets i^\star\Delta$
\Else
  \State Try fallback $\mathcal{P}^{\mathrm{fb}}$ (analytic or $\pi_b^{\mathrm{NN}}$); update $\mathcal{P}_{\mathrm{buf}}, T_k$ if certified, else $\mathrm{commit}\gets\mathrm{false}$
\EndIf
\State Execute next step of $\mathcal{P}_{\mathrm{buf}}$ and advance index
\For{each open record $(j, k')$ with $(k-k')\Delta \le T_{k'}$}
  \State Update $\hat s_{jk'}$ at lag $t=(k-k')\Delta$ via~\eqref{eq:sensed}
\EndFor
\For{each class $c$ with records reaching $k-k' = k^\star_{k'}$}
  \State Compute finalized Tracker Miss $\hat m_c$; update $q_c$ by~\eqref{eq:update}
\EndFor
\If{$\mathrm{commit}$}
  \State Open record $(j, k)$ with $(c(j), q_{c(j)}, T_k)$ for each retained $j$
\EndIf
\end{algorithmic}
\end{algorithm}

\subsection{Certificate-Aligned Learned Backup Policy \texorpdfstring{$\pi_b^{\mathrm{NN}}$}{pi\_b\^{}NN}}
Under non-yielding $\Sterm$, analytic backups stall outside $\mathcal{F}_{\mathrm{traffic}}$.
We train a causal learned backup $\pi_b^{\mathrm{NN}}$ directly against the same dynamics, error tubes, and exposure-window certificate used online, without querying future obstacle trajectories.
Every learned rollout executes a bounded-duration active manoeuvre followed by an exact braking tail into $\Sterm$, ensuring clean alignment with calibrated braking error tubes.

\textbf{Causal observations and manoeuvre parameterization.}
$\pi_b^{\mathrm{NN}}$ observes only causal features $o_k$---ego state, static clearances, the $K_{\mathrm{ret}}$ nearest sensed obstacles with their live bounds $q_c$ and multi-lag capsule margins, the static envelope $\mathcal{F}_{\mathrm{traffic}}$, and $u^{\mathrm{nom}}_k$.
To avoid mode collapse across corridors and side bays, $\pi_b^{\mathrm{NN}}$ outputs $M$ complementary finite-duration active modes $m \in \{1,\ldots,M\}$ (forward progress, lateral avoidance or refuge entry/exit, and high-margin short-$t^\star$ retreat), each switching cleanly to an exact braking tail into $\Sterm$:
\begin{equation}
u^{(m)}_i = \begin{cases}
\pi_{b,\theta}^{(m)}(o_k, i), & 0 \le i < \tau_m, \\
u_{\mathrm{brake}}(\hat x_i), & \tau_m \le i < N,
\end{cases}
\label{eq:backup_param}
\end{equation}
where $\tau_m \in [\tau_{\min}, \tau_{\max}]$, $\|u^{(m)}_i\| \le u_{\max}$, and $u_{\mathrm{brake}}$ brings the rollout into $\Sterm$.

\textbf{Two-stage certificate-aligned training.}
Let $\xi^{(m)}$ denote the prediction-model rollout under~\eqref{eq:backup_param}, $h_{\mathrm{cert}}(\xi^{(m)}; q)$ its minimum signed clearance across conditions~(i)--(ii) (positive iff certified into $\Sterm$), $t^\star(\xi^{(m)})$ its entry time, and $g(\xi^{(m)})$ its goal progress.
We train $\theta$ in two stages over randomized per-class bounds $q_c \in [0.05, 1.0]$, selecting the final policy on validation states.
Stage~1 warm-starts $\theta$ by imitating certified demonstration manoeuvres collected offline under the exposure-window certificate to establish mode diversity.
Stage~2 minimizes the unified certificate-aligned objective
\begin{equation}
\begin{aligned}
\mathcal{L}(\theta) = \mathbb{E}_{o, q}\!\big[
&\lambda_s [m_0 - h_{\mathrm{cert}}(\xi; q)]_+
- \lambda_g \mathbb{I}_{\mathrm{cert}}\, g(\xi) \\
&+ \lambda_t\, t^\star(\xi)
+ \mathcal{R}_{\mathrm{reg}}(\theta)\big],
\end{aligned}
\label{eq:backup_loss}
\end{equation}
where $[z]_+ {=} \max(0,z)$ enforces a safety-margin hinge above $m_0{>}0$, $\mathbb{I}_{\mathrm{cert}} {=} \mathbb{I}[h_{\mathrm{cert}}(\xi; q) \ge 0]$ gates progress strictly on certified rollouts (aggregated over modes $m$), $\lambda_t t^\star(\xi)$ encourages brief exposure windows, and $\mathcal{R}_{\mathrm{reg}}(\theta)$ combines action-smoothness regularization with a light Stage-1 demonstration anchor.
Online, Algorithm~\ref{alg:unified} evaluates all candidate modes in parallel for both nominal continuations and direct fallbacks, committing the highest-progress certified nominal plan when available and falling back to the best certified direct backup otherwise (including overriding a stalled nominal plan when a direct backup yields substantial certified progress).

\subsection{Coverage Bound and Replanning Safety Hierarchy}
\begin{proposition}[Long-run Tracker Miss bound]\label{prop:aci}
Let $t = 1,\ldots,T$ index updates of class $c$ via~\eqref{eq:update} at $q_{\max}{=}1$ with Tracker Miss $\hat m_{c,t}$, and let $U$ be the number of upper clips at $q_c{=}1$. For any score sequence and causal finalization rule,
\[
\frac{1}{T}\sum_{t=1}^{T}\hat m_{c,t} \;\le\; \alpha + \frac{1 - q_{c,1}}{\eta T} + \frac{U}{T}.
\]
\end{proposition}
\begin{proof}
Write $q_{t+1} = q_t + \eta(\hat m_t - \alpha) - u_t + l_t$ with $u_t,l_t \ge 0$. Summing over $t=1,\ldots,T$ gives $\eta\sum_t(\hat m_t - \alpha) = q_{T+1} - q_1 + \sum_t u_t - \sum_t l_t \le 1 - q_1 + \eta U$ since $q_{T+1}\le 1$ and $u_t \le \eta$.
\end{proof}

\begin{table*}[t]
\centering
\caption{Experimental settings across terminal-set conditions and simulators.}\label{tab:exp_setup}
\scriptsize\setlength{\tabcolsep}{4.0pt}
\resizebox{\textwidth}{!}{%
\begin{tabular}{@{}llcccc@{}}
\toprule
Condition (Terminal Set) & Platform (Dynamics) & Scenarios & $N$ ($H$, $\Delta$) & Tracker $(\alpha, \eta)$ & Sensing / Mismatch \\
\midrule
\textbf{Regime~I} ($\SH$, Yielding) & Isaac Sim 3D (Quadrotor) & 5 (\emph{Crowd, Corridor, Open, Pursuit, Cross}) & 136 (2.19\,s, 1/62\,s) & $(0.0025, 0.05)$ & 36-ray LiDAR, $(e_p,e_v,e_z)$ \\
\textbf{Regime~II} ($\Sterm$, Non-Yielding) & Isaac Sim 3D (Quadrotor) & 4 (\emph{Side Pocket, Crossing, Stop-Go, Motion Shift}) & 136 (2.19\,s, 1/62\,s) & $(0.01, 0.50)$ & 36-ray LiDAR, $(e_p,e_v,e_z)$ \\
\midrule
\textbf{Regime~I} ($\SH$, Yielding) & Planar DI 2D & 9 (\emph{Open, Side Exit, Crossing, Turning, Stop-Go,} \ldots) & 35 (3.50\,s, 0.1\,s) & $(0.01, 0.25)$ & Noisy obs., exact 2D \\
\textbf{Regime~II} ($\Sterm$, Non-Yielding) & Planar DI 2D & 9 (same 9 scenarios without $A_h$) & 35 (3.50\,s, 0.1\,s) & $(0.01, 0.25)$ & Noisy obs., exact 2D \\
\bottomrule
\end{tabular}%
}\\[1pt]
{\footnotesize All certified runs use $W{=}N$, $K_{\mathrm{ret}}\le 5$, $q_c\in[0,1]$, and a 30\,s episode horizon on held-out evaluation trials.}
\vspace{-4pt}
\end{table*}

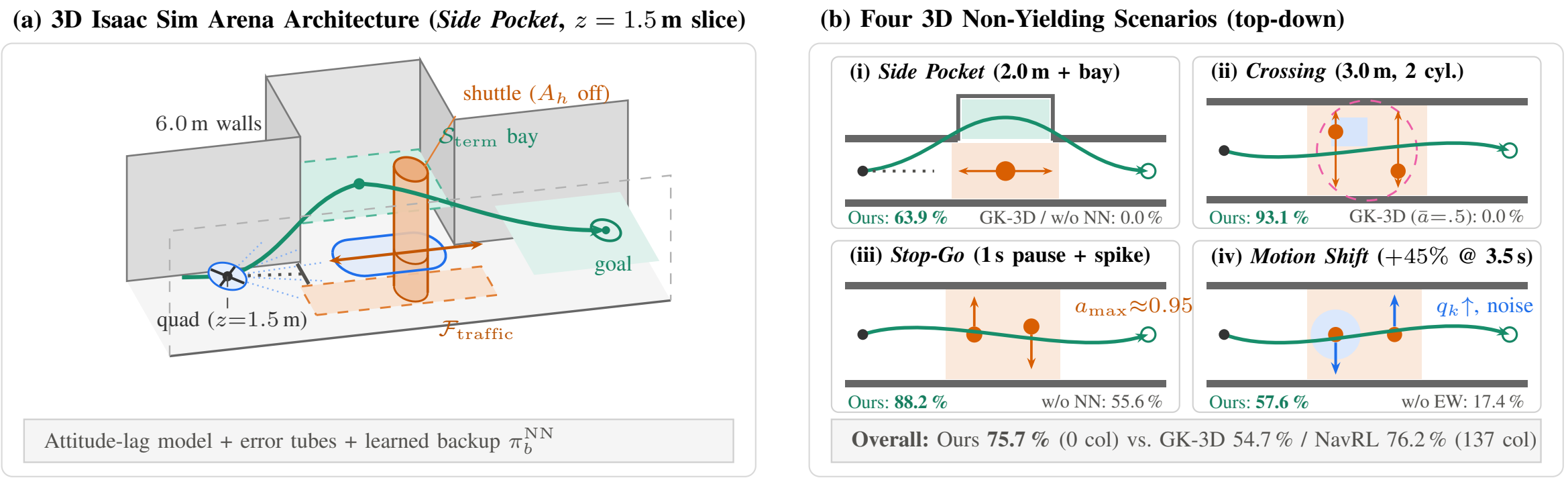
\begin{figure*}[t]
\centering
\begin{tikzpicture}[x=1cm,y=1cm,font=\footnotesize,scale=0.88,transform shape]
  \definecolor{fastc}{HTML}{1F6FEB}
  \definecolor{slowc}{HTML}{D95F02}
  \definecolor{termc}{HTML}{1B9E77}
  \definecolor{lanec}{HTML}{E7298A}
  \definecolor{ink2}{HTML}{52514E}
  \definecolor{wallc}{HTML}{333333}

  \begin{scope}[shift={(0,0)}]
    \node[anchor=west, font=\footnotesize\bfseries] at (0.0, 3.80) {(a) 3D Isaac Sim Arena Architecture (\emph{Side Pocket}, $z=1.5$\,m slice)};
    \draw[black!15, rounded corners=3pt] (0.0, -1.30) rectangle (8.45, 3.55);

    \begin{scope}[shift={(4.45,0.78)}, x={(0.54cm,0.06cm)}, y={(-0.24cm,0.42cm)}, z={(0cm,0.28cm)}]
      \fill[black!4] (-5.2,-1.0,0) -- (5.2,-1.0,0) -- (5.2,1.0,0) -- (1.6,1.0,0) -- (1.6,2.6,0) -- (-1.6,2.6,0) -- (-1.6,1.0,0) -- (-5.2,1.0,0) -- cycle;
      \fill[slowc!18] (-1.88,-0.38,0) -- (1.88,-0.38,0) -- (1.88,0.38,0) -- (-1.88,0.38,0) -- cycle;
      \draw[slowc!80, dashed, line width=0.55pt] (-1.88,-0.38,0) -- (1.88,-0.38,0) -- (1.88,0.38,0) -- (-1.88,0.38,0) -- cycle;

      \fill[wallc!13, draw=wallc!65, line width=0.5pt] (-1.6,2.6,0) -- (1.6,2.6,0) -- (1.6,2.6,5.0) -- (-1.6,2.6,5.0) -- cycle;
      \fill[wallc!22, draw=wallc!65, line width=0.5pt] (-1.6,1.0,0) -- (-1.6,2.6,0) -- (-1.6,2.6,5.0) -- (-1.6,1.0,5.0) -- cycle;
      \fill[wallc!24, draw=wallc!65, line width=0.5pt] (1.6,1.0,0) -- (1.6,2.6,0) -- (1.6,2.6,5.0) -- (1.6,1.0,5.0) -- cycle;
      \fill[termc!18] (-1.55,1.02,1.5) -- (1.55,1.02,1.5) -- (1.55,2.55,1.5) -- (-1.55,2.55,1.5) -- cycle;
      \draw[termc!75, dashed, line width=0.55pt] (-1.55,1.02,1.5) -- (1.55,1.02,1.5) -- (1.55,2.55,1.5) -- (-1.55,2.55,1.5) -- cycle;
      \fill[wallc!17, draw=wallc!65, line width=0.5pt] (-5.2,1.0,0) -- (-1.6,1.0,0) -- (-1.6,1.0,5.0) -- (-5.2,1.0,5.0) -- cycle;
      \fill[wallc!17, draw=wallc!65, line width=0.5pt] (1.6,1.0,0) -- (5.2,1.0,0) -- (5.2,1.0,5.0) -- (1.6,1.0,5.0) -- cycle;

      \fill[termc!14] (2.4, -0.92, 1.5) -- (5.0, -0.92, 1.5) -- (5.0, 0.92, 1.5) -- (2.4, 0.92, 1.5) -- cycle;

      \fill[fastc!15, fill opacity=0.68]
        plot[domain=-90:90, samples=18] ({0.45 + 0.48*cos(\x)}, {0.0 + 0.48*sin(\x)}, 1.5) --
        plot[domain=90:270, samples=18] ({-0.85 + 0.48*cos(\x)}, {0.0 + 0.48*sin(\x)}, 1.5) -- cycle;
      \draw[fastc, line width=0.7pt]
        plot[domain=-24:90, samples=14] ({0.45 + 0.48*cos(\x)}, {0.0 + 0.48*sin(\x)}, 1.5) --
        plot[domain=90:156, samples=10] ({-0.85 + 0.48*cos(\x)}, {0.0 + 0.48*sin(\x)}, 1.5);

      \fill[slowc!28] plot[domain=0:360, samples=32, smooth cycle] ({0.25 + 0.32*cos(\x)}, {0.0 + 0.32*sin(\x)}, 0);
      \fill[slowc!36, opacity=0.90]
        plot[domain=156:336, samples=18] ({0.25 + 0.32*cos(\x)}, {0.0 + 0.32*sin(\x)}, 0) --
        ({0.25 + 0.32*cos(336)}, {0.0 + 0.32*sin(336)}, 4.8) --
        plot[domain=336:156, samples=18] ({0.25 + 0.32*cos(\x)}, {0.0 + 0.32*sin(\x)}, 4.8) -- cycle;
      \draw[slowc!90!black, line width=0.7pt]
        ({0.25 + 0.32*cos(156)}, {0.0 + 0.32*sin(156)}, 0) -- ({0.25 + 0.32*cos(156)}, {0.0 + 0.32*sin(156)}, 4.8)
        ({0.25 + 0.32*cos(336)}, {0.0 + 0.32*sin(336)}, 0) -- ({0.25 + 0.32*cos(336)}, {0.0 + 0.32*sin(336)}, 4.8);
      \draw[slowc!90!black, line width=0.55pt] plot[domain=156:336, samples=18] ({0.25 + 0.32*cos(\x)}, {0.0 + 0.32*sin(\x)}, 0);
      \draw[slowc!85!black, line width=0.55pt] plot[domain=156:336, samples=18] ({0.25 + 0.32*cos(\x)}, {0.0 + 0.32*sin(\x)}, 1.5);
      \fill[slowc!50, draw=slowc!90!black, line width=0.65pt]
        plot[domain=0:360, samples=32, smooth cycle] ({0.25 + 0.32*cos(\x)}, {0.0 + 0.32*sin(\x)}, 4.8);

      \draw[fastc, line width=0.7pt]
        plot[domain=156:270, samples=14] ({-0.85 + 0.48*cos(\x)}, {0.0 + 0.48*sin(\x)}, 1.5) --
        plot[domain=-90:-24, samples=10] ({0.45 + 0.48*cos(\x)}, {0.0 + 0.48*sin(\x)}, 1.5);
      \draw[{Stealth[length=3.5pt]}-{Stealth[length=3.5pt]}, slowc!90!black, line width=0.8pt] (-1.45,0.0,1.5) -- (1.75,0.0,1.5);

      \coordinate (drone3d) at (-3.6, -0.1, 1.5);
      \coordinate (pocket3d) at (0.0, 1.85, 1.5);
      \coordinate (goal3d) at (4.3, 0.0, 1.5);
      \coordinate (start3d) at (-4.5, 0.0, 1.5);
      \foreach \ang in {-55,-25,5,32,58} {
        \draw[fastc!55, densely dotted, line width=0.45pt] (drone3d) -- ++({2.0*cos(\ang)}, {0.90*sin(\ang)}, 0);
      }
      \draw[termc!90!black, line width=1.15pt, -{Stealth[length=4.5pt]}]
        (start3d) -- (drone3d) .. controls (-2.2, 0.2, 1.5) and (-1.3, 1.65, 1.5) .. (pocket3d) .. controls (1.2, 1.65, 1.5) and (2.2, 0.3, 1.5) .. (goal3d);
      \fill[termc!90!black] (pocket3d) circle (1.8pt);
      \draw[ink2, dotted, line width=0.85pt] (drone3d) -- (-2.1, -0.25, 1.5);
      \draw[ink2, line width=1.0pt] (-2.1, -0.52, 1.5) -- (-2.1, 0.02, 1.5);

      \fill[fastc!12, draw=fastc, line width=0.55pt]
        plot[domain=0:360, samples=32, smooth cycle] ({-3.6 + 0.36*cos(\x)}, {-0.1 + 0.36*sin(\x)}, 1.5);
      \draw[wallc, line width=0.75pt] (-3.85,-0.32,1.5) -- (-3.35,0.12,1.5) (-3.85,0.12,1.5) -- (-3.35,-0.32,1.5);
      \fill[wallc] (drone3d) circle (1.6pt);
      \draw[ink2, dashed, line width=0.4pt] (drone3d) -- (-3.6, -0.1, 0);
      \draw[termc!85!black, line width=0.75pt]
        plot[domain=0:360, samples=32, smooth cycle] ({4.3 + 0.28*cos(\x)}, {0.0 + 0.28*sin(\x)}, 1.5);
      \fill[termc!85!black] (goal3d) circle (1.3pt);
      \draw[wallc!75, line width=0.65pt] (-5.2,-1.0,0) -- (5.2,-1.0,0);
      \draw[wallc!38, dashed, line width=0.45pt] (-5.2,-1.0,0) -- (-5.2,-1.0,5.0) -- (5.2,-1.0,5.0) -- (5.2,-1.0,0);

      \node[anchor=south, font=\scriptsize, wallc] at (-3.5, 1.0, 5.20) {$6.0$\,m walls};
      \node[anchor=south west, font=\scriptsize, slowc!90!black] at (1.55, 1.0, 5.20) {shuttle ($A_h$ off)};
      \draw[slowc!80, line width=0.45pt] (1.60, 0.95, 5.22) -- (0.55, 0.05, 4.75);
      \node[anchor=west, font=\scriptsize, termc!85!black] at (1.55, 2.15, 2.55) {$\mathcal{S}_{\mathrm{term}}$ bay};
      \node[anchor=north west, font=\scriptsize, slowc!85!black] at (0.35, -0.55, 0.0) {$\mathcal{F}_{\mathrm{traffic}}$};
      \node[anchor=north, font=\scriptsize, wallc] at (-3.7, -0.55, 1.25) {quad ($z{=}1.5$\,m)};
      \node[anchor=north, font=\scriptsize, termc!85!black] at (4.3, -0.35, 1.45) {goal};
    \end{scope}

    \fill[black!4] (0.25, -1.14) rectangle (8.20, -0.66);
    \draw[black!18, line width=0.45pt] (0.25, -1.14) rectangle (8.20, -0.66);
    \node[anchor=west, font=\scriptsize, ink2] at (0.35, -0.90) {Attitude-lag model + error tubes + learned backup $\pi_b^{\mathrm{NN}}$};
  \end{scope}

  \begin{scope}[shift={(9.05,0)}]
    \node[anchor=west, font=\footnotesize\bfseries] at (0.0, 3.80) {(b) Four 3D Non-Yielding Scenarios (top-down)};
    \draw[black!15, rounded corners=3pt] (0.0, -1.30) rectangle (8.45, 3.55);

    \begin{scope}[shift={(0.25, 1.48)}]
      \draw[black!18, rounded corners=2pt] (0,0) rectangle (3.90, 1.92);
      \node[anchor=west, font=\scriptsize\bfseries] at (0.08, 1.74) {(i) \emph{Side Pocket} (2.0\,m + bay)};
      \fill[wallc!75] (0.15, 0.96) rectangle (3.75, 1.04);
      \fill[wallc!75] (0.15, 0.24) rectangle (3.75, 0.32);
      \fill[white] (1.42, 0.96) rectangle (2.48, 1.48);
      \draw[wallc!75, line width=1.1pt] (1.42, 0.98) -- (1.42, 1.48) -- (2.48, 1.48) -- (2.48, 0.98);
      \fill[termc!18] (1.46, 1.00) rectangle (2.44, 1.44);
      \fill[slowc!16] (1.35, 0.33) rectangle (2.55, 0.95);
      \fill[slowc] (1.95, 0.64) circle (0.11);
      \draw[{Stealth[length=2.8pt]}-{Stealth[length=2.8pt]}, slowc, line width=0.6pt] (1.42, 0.64) -- (2.48, 0.64);
      \draw[termc!90!black, line width=1.0pt, -{Stealth[length=3.5pt]}] (0.35, 0.64) .. controls (1.10, 0.72) and (1.38, 1.24) .. (1.95, 1.24) .. controls (2.52, 1.24) and (2.80, 0.70) .. (3.55, 0.64);
      \draw[ink2, dotted, line width=0.8pt] (0.35, 0.64) -- (1.15, 0.64);
      \fill[wallc] (0.35, 0.64) circle (0.06);
      \draw[termc!85!black, line width=0.6pt] (3.55, 0.64) circle (0.08);
      \node[anchor=west, font=\fontsize{6}{7}\selectfont, termc!80!black] at (0.06, 0.12) {Ours: \textbf{63.9\,\%}};
      \node[anchor=east, font=\fontsize{6}{7}\selectfont, ink2] at (3.84, 0.12) {GK-3D / w/o NN: 0.0\,\%};
    \end{scope}

    \begin{scope}[shift={(4.30, 1.48)}]
      \draw[black!18, rounded corners=2pt] (0,0) rectangle (3.90, 1.92);
      \node[anchor=west, font=\scriptsize\bfseries] at (0.08, 1.74) {(ii) \emph{Crossing} (3.0\,m, 2 cyl.)};
      \fill[wallc!75] (0.15, 1.38) rectangle (3.75, 1.46);
      \fill[wallc!75] (0.15, 0.28) rectangle (3.75, 0.36);
      \fill[slowc!14] (1.28, 0.37) rectangle (2.62, 1.37);
      \draw[lanec!75, dashed, line width=0.55pt] (1.95, 0.87) circle (0.56);
      \fill[fastc!18] (1.55, 0.92) rectangle (1.95, 1.24);
      \fill[slowc] (1.60, 1.08) circle (0.085);
      \fill[slowc] (2.30, 0.64) circle (0.085);
      \draw[{Stealth[length=2.8pt]}-{Stealth[length=2.8pt]}, slowc, line width=0.55pt] (1.60, 0.45) -- (1.60, 1.32);
      \draw[{Stealth[length=2.8pt]}-{Stealth[length=2.8pt]}, slowc, line width=0.55pt] (2.30, 0.45) -- (2.30, 1.32);
      \draw[termc!90!black, line width=1.0pt, -{Stealth[length=3.5pt]}] (0.35, 0.87) .. controls (1.15, 0.62) and (2.65, 1.12) .. (3.55, 0.87);
      \fill[wallc] (0.35, 0.87) circle (0.06);
      \draw[termc!85!black, line width=0.6pt] (3.55, 0.87) circle (0.08);
      \node[anchor=west, font=\fontsize{6}{7}\selectfont, termc!80!black] at (0.06, 0.12) {Ours: \textbf{93.1\,\%}};
      \node[anchor=east, font=\fontsize{6}{7}\selectfont, ink2] at (3.84, 0.12) {GK-3D ($\abar{=}.5$): 0.0\,\%};
    \end{scope}

    \begin{scope}[shift={(0.25, -0.58)}]
      \draw[black!18, rounded corners=2pt] (0,0) rectangle (3.90, 1.92);
      \node[anchor=west, font=\scriptsize\bfseries] at (0.08, 1.74) {(iii) \emph{Stop-Go} (1\,s pause + spike)};
      \fill[wallc!75] (0.15, 1.38) rectangle (3.75, 1.46);
      \fill[wallc!75] (0.15, 0.28) rectangle (3.75, 0.36);
      \fill[slowc!14] (1.28, 0.37) rectangle (2.56, 1.37);
      \fill[slowc] (1.60, 0.87) circle (0.09);
      \fill[slowc] (2.24, 0.96) circle (0.09);
      \draw[slowc, line width=0.7pt, -{Stealth[length=3.2pt]}] (1.60, 0.96) -- (1.60, 1.30);
      \draw[slowc, line width=0.7pt, -{Stealth[length=3.2pt]}] (2.24, 0.87) -- (2.24, 0.48);
      \node[slowc!90!black, font=\scriptsize, anchor=west] at (2.60, 1.18) {$a_{\max}{\approx}0.95$};
      \draw[termc!90!black, line width=1.0pt, -{Stealth[length=3.5pt]}] (0.35, 0.87) .. controls (1.15, 1.14) and (2.65, 0.60) .. (3.55, 0.87);
      \fill[wallc] (0.35, 0.87) circle (0.06);
      \draw[termc!85!black, line width=0.6pt] (3.55, 0.87) circle (0.08);
      \node[anchor=west, font=\fontsize{6}{7}\selectfont, termc!80!black] at (0.06, 0.12) {Ours: \textbf{88.2\,\%}};
      \node[anchor=east, font=\fontsize{6}{7}\selectfont, ink2] at (3.84, 0.12) {w/o NN: 55.6\,\%};
    \end{scope}

    \begin{scope}[shift={(4.30, -0.58)}]
      \draw[black!18, rounded corners=2pt] (0,0) rectangle (3.90, 1.92);
      \node[anchor=west, font=\scriptsize\bfseries] at (0.08, 1.74) {(iv) \emph{Motion Shift} ($+45\%$ @ 3.5\,s)};
      \fill[wallc!75] (0.15, 1.38) rectangle (3.75, 1.46);
      \fill[wallc!75] (0.15, 0.28) rectangle (3.75, 0.36);
      \fill[slowc!14] (1.28, 0.37) rectangle (2.56, 1.37);
      \fill[fastc!16] (1.60, 0.87) circle (0.28);
      \fill[slowc] (1.60, 0.87) circle (0.085);
      \fill[slowc] (2.26, 0.87) circle (0.085);
      \draw[fastc, line width=0.75pt, -{Stealth[length=3.5pt]}] (1.60, 0.78) -- (1.60, 0.42);
      \draw[fastc, line width=0.75pt, -{Stealth[length=3.5pt]}] (2.26, 0.96) -- (2.26, 1.32);
      \node[fastc, font=\scriptsize, anchor=west] at (2.60, 1.18) {$q_k{\uparrow}$, noise};
      \draw[termc!90!black, line width=1.0pt, -{Stealth[length=3.5pt]}] (0.35, 0.87) .. controls (1.20, 0.56) and (2.60, 1.16) .. (3.55, 0.87);
      \fill[wallc] (0.35, 0.87) circle (0.06);
      \draw[termc!85!black, line width=0.6pt] (3.55, 0.87) circle (0.08);
      \node[anchor=west, font=\fontsize{6}{7}\selectfont, termc!80!black] at (0.06, 0.12) {Ours: \textbf{57.6\,\%}};
      \node[anchor=east, font=\fontsize{6}{7}\selectfont, ink2] at (3.84, 0.12) {w/o EW: 17.4\,\%};
    \end{scope}

    \fill[black!4] (0.25, -1.14) rectangle (8.20, -0.66);
    \draw[black!18, line width=0.45pt] (0.25, -1.14) rectangle (8.20, -0.66);
    \node[anchor=west, font=\scriptsize, ink2] at (0.35, -0.90) {\textbf{Overall:} Ours \textbf{75.7\,\%} (0 col) vs.\ GK-3D 54.7\,\% / NavRL 76.2\,\% (137 col)};
  \end{scope}
\end{tikzpicture}
\vspace{-4pt}
\caption{Isaac Sim 3D non-yielding ($\Sterm$) benchmark geometry: (a) isometric \emph{Side Pocket} with $\mathcal{F}_{\mathrm{traffic}}$ and lateral refuge bay $\Sterm$; (b) top-down layouts and per-scene safe completion rates across all four scenarios.}\label{fig:isaac_3d_schematic}
\vspace{-4pt}
\end{figure*}

Proposition~\ref{prop:aci} holds without exchangeability under performative shifts~\cite{perdomo2020performative}, with Plan Miss linked to $\alpha$ via the union bound across retained obstacles.
Specifically, Lemma~\ref{lem:exposure} provides conditional deterministic safety when obstacles remain inside their certified capsules on $[0, t^\star]$, whereas Proposition~\ref{prop:aci} degrades explicitly by $U/T$ whenever obstacle accelerations exceed $q_{\max}{=}1$.
In closed-loop execution, however, Algorithm~\ref{alg:unified} re-commits a fresh plan after $k_{\mathrm{exec}} \ll k^\star$ steps, typically after a single control step.
A breach occurring strictly at lag $t > k_{\mathrm{exec}}\Delta$ affects only a superseded tail, whereas physical collision requires a ground-truth breach on an executed step---which we define as an \emph{Execute Miss}---yielding the nested hierarchy
\begin{equation}
P(\text{collision}) \;\le\; P(\text{Execute Miss}) \;\le\; P(\text{Plan Miss}),
\label{eq:hierarchy}
\end{equation}
which we evaluate alongside $U/T$ in Sec.~\ref{sec:exp_calib}.
Tracker overhead scales as $O(WK)$ for $K \le 5$, keeping real-time execution well within the control period.

\section{Experiments}\label{sec:experiments}

\textbf{Experimental conditions and platforms.}
Every certified run wraps the frozen NavRL nominal policy~\cite{xu2025navrl} with the same online per-class tracker (entry-step finalization at $k^\star$, $W{=}N$, $q_c \in [0,1]$, worst-case bound $\abar{=}1$\,m/s$^2$).
Table~\ref{tab:exp_setup} lists the test setups across both terminal-set conditions ($\SH$ vs.\ $\Sterm$) and both simulators: a 3D quadrotor in Isaac Sim using calibrated four-state attitude-lag rollouts, mismatch tubes $(e_p,e_v,e_z)$, 36-ray 4\,m LiDAR, and $\delta_\Delta{=}0.038$\,m, alongside a planar 2D double integrator.
We examine worst-case bottlenecks (Sec.~\ref{sec:exp_bottleneck}), tracker ablations (Sec.~\ref{sec:exp_calib}), and calibration--backup coupling (Sec.~\ref{sec:exp_coupling}) in turn, before comparing \emph{Ours} with certified and reactive baselines across all four combinations of terminal-set condition and simulator in Sec.~\ref{sec:exp_baselines}.

\subsection{Performance Bottlenecks Under Two Terminal-Set Conditions}\label{sec:exp_bottleneck}

Before introducing calibration or comparing external baselines, we first ask whether conservatism under $\SH$ and $\Sterm$ comes from the same bottleneck.
To avoid confounding dynamics or algorithm differences, Table~\ref{tab:bottleneck_3d} keeps the simulator, nominal policy, and worst-case bound ($q{=}1$) identical in Isaac Sim 3D, varying only the terminal-set condition and the backup/window choice (Fig.~\ref{fig:isaac_3d_schematic}).

\begin{table}[t]
\centering
\caption{Worst-case occupancy ($q{=}1$) performance in Isaac Sim 3D.}\label{tab:bottleneck_3d}
\scriptsize\setlength{\tabcolsep}{3.0pt}
\resizebox{\columnwidth}{!}{%
\begin{tabular}{@{}lcccccc@{}}
\toprule
& \multicolumn{3}{c}{\textbf{Regime~I} ($\SH$, Crowd)} & \multicolumn{3}{c}{\textbf{Regime~II} ($\Sterm$, 4 scenes)} \\
\cmidrule(lr){2-4}\cmidrule(lr){5-7}
Filter Setting & Comp.\ (\%) & Col. & $t^\star$ (s) & Comp.\ (\%) & Col. & $t^\star$ (s) \\
\midrule
NavRL (Unfiltered) & 95.5 & 39 & --- & 76.22 & 137 & --- \\
Worst-Case ($q{=}1$, Full-$H$, Analytic) & 83.1 & 0 & 1.24 & 0.00 & 0 & 2.193 \\
Worst-Case ($q{=}1$, EW, Analytic) & 83.1 & 0 & 1.24 & 0.00 & 0 & 0.868 \\
Worst-Case ($q{=}1$, EW, Learned) & 83.4 & 0 & 1.24 & 0.69 & 0 & 1.055 \\
\bottomrule
\end{tabular}%
}\\[1pt]
{\footnotesize Same Isaac Sim 3D filter architecture with $q{=}1$ fixed; only the terminal-set condition ($\SH$ vs.\ $\Sterm$) and backup/window switch.}
\vspace{-4pt}
\end{table}

Under passive yielding ($\SH$), any collision-free braking trajectory ends inside the terminal set.
Because a quadrotor's stopping time from cruise speed is governed largely by thrust and attitude limits ($t^\star \approx 1.24$\,s), switching from an analytic backup to a learned one neither shortens $t^\star$ nor displaces far enough laterally within one second to clear a worst-case $q{=}1$ capsule.
Consequently, every tested backup loses roughly $12$\,pp of Crowd completion relative to unfiltered NavRL (spread within $1.1$\,pp across backups), indicating that worst-case occupancy growth over $[0, t^\star]$ is the sole active bottleneck.

Dropping yielding under $\Sterm$ alters what a certified backup must physically achieve: hovering inside a traffic lane is no longer safe, so every feasible rollout must cross or exit $\mathcal{F}_{\mathrm{traffic}}$ into $\Sterm$.
Under full-horizon certification (\emph{Full-$H$}), quadratic capsule growth $\tfrac12 \abar t^2$ at $q{=}1$ spans the entire 2--3\,m corridor width and blocks entry completely ($0.00\,\%$).
Even when verification is truncated to $[0, t^\star]$ (\emph{EW}) or paired with a learned backup, uncalibrated $q{=}1$ capsules still leave too little clearance to enter traffic ($\le 0.69\,\%$).
Unlike under $\SH$, neither window truncation nor backup learning can overcome worst-case occupancy inflation on its own.

\subsection{Online Exposure-Window Calibration and Internal Ablations}\label{sec:exp_calib}

\begin{figure*}[t]
\centering
\includegraphics[width=0.92\textwidth]{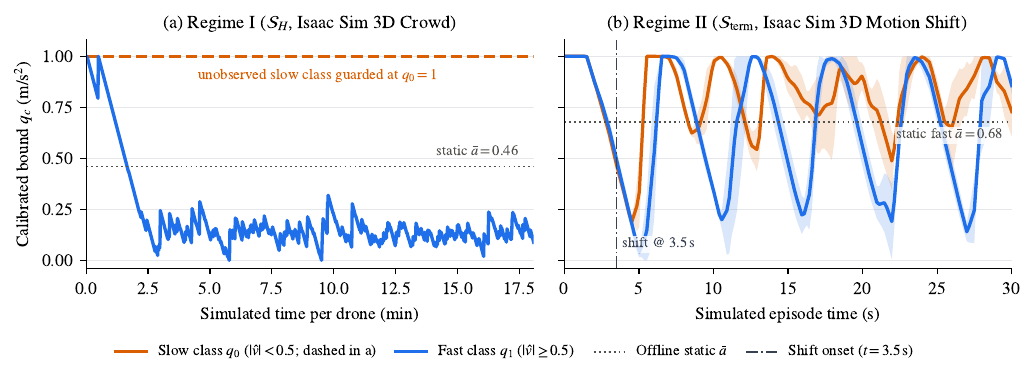}
\vspace{-4pt}
\caption{Isaac Sim 3D online per-class calibration dynamics: (a) Regime~I ($\SH$, Crowd), where $q_{\mathrm{fast}}$ settles within three simulated minutes while unobserved slow obstacles stay guarded at $q_{\mathrm{slow}}{=}1$; (b) Regime~II ($\Sterm$, \emph{Motion Shift}), where both class bounds spike after the $t{=}3.5$\,s shift while $q_{\mathrm{fast}}$ relaxes below the offline static bound ($0.68$) between reversals, allowing corridor passage.}\label{fig:q_traj_3d}
\vspace{-4pt}
\end{figure*}

Because fixing $q{=}1$ stalls navigation, we next test whether online adaptation, speed-class separation, and exposure-window truncation are each necessary, or whether a simpler static or pooled bound suffices (Table~\ref{tab:calib_ablation_3d}, Fig.~\ref{fig:q_traj_3d}).

\begin{table*}[t]
\centering
\caption{Online calibration validation and internal ablations in Isaac Sim 3D (zero collisions across all certified methods).}\label{tab:calib_ablation_3d}
\scriptsize\setlength{\tabcolsep}{3.4pt}
\resizebox{\textwidth}{!}{%
\begin{tabular}{@{}lcccccccccc@{}}
\toprule
& \multicolumn{4}{c}{\textbf{Regime~I} ($\SH$, Isaac Sim 3D)} & \multicolumn{6}{c}{\textbf{Regime~II} ($\Sterm$, Isaac Sim 3D, 4 scenes)} \\
\cmidrule(lr){2-5}\cmidrule(lr){6-11}
Calibration Variant & Crowd (\%) & Corr.\ (\%) & Purs.\ (\%) & Max Plan Miss & Overall (\%$\pm$SD) & Side Pkt. & Cross. & $t^\star$ (s) & Tracker / Plan Miss & Execute Miss \\
\midrule
\textbf{Ours (Online, Per-Class, EW)} & \textbf{95.6} & \textbf{20.4} & \textbf{52.0} & \textbf{0.995\,\%} & \textbf{75.69 $\pm$ 1.20} & \textbf{63.9} & \textbf{93.1} & 1.512 & 10.02 / 17.02\,\% & 0.093\,\% \\
Ours w/o EW (Full-$H$ $[0,H]$) & 95.6$^\dagger$ & 20.4$^\dagger$ & 52.0$^\dagger$ & 0.995\,\% & 48.44 $\pm$ 1.38 & 16.0 & 79.2 & 2.194 & 24.30 / 33.29\,\% & 0.283\,\% \\
Ours (Pooled Single $q$) & 94.6 & 19.3 & --- & \textbf{59.0\,\%$^*$} & 67.53 $\pm$ 2.57 & 53.5 & 91.7 & 1.471 & 10.88 / 17.01\,\% & 0.073\,\% \\
Ours (Static Calib.) & 91.5 & 21.1 & 45.5 & 2.92\,\% & 13.19 $\pm$ 2.41 & 52.8 & 0.0 & 1.176 & 0.00 / 0.52\,\% & 0.006\,\% \\
Ours w/o Calib.\ ($q{=}1$) & 83.1 & 18.1 & 48.0 & 0.00\,\% & 0.69 $\pm$ 0.30 & 2.8 & 0.0 & 1.055 & 0.00 / 0.31\,\% & 0.002\,\% \\
\bottomrule
\end{tabular}%
}\\[1pt]
{\footnotesize $^\dagger$Under $\SH$, braking fixes $t^\star$ and yields identical passive-safety horizon; $^*$slow-class miss rate in Crowd (Corridor slow-class miss reaches $2.99\,\%$ vs.\ $0.715\,\%$ per-class). Static Calib.\ under $\Sterm$ is tuned offline on a separate validation set.}
\vspace{-4pt}
\end{table*}

An offline bound must pick a single compromise across all traffic regimes.
Under $\SH$, that compromise already lags online tracking in Crowd ($95.6\,\%$ vs.\ $91.5\,\%$) while incurring higher Plan Miss in Pursuit (Table~\ref{tab:calib_ablation_3d}).
Under $\Sterm$, the failure becomes acute: reversal and acceleration spikes on the offline validation set force the static bound upward to $q_{\mathrm{static}}{=}(0.96, 0.68)$, freezing corridor entry in all three crossing scenes ($13.19\,\%$ overall vs.\ $75.69\,\%$ online).
Fig.~\ref{fig:q_traj_3d}(b) reveals why online tracking succeeds: $q_{\mathrm{fast}}$ jumps immediately after the $t{=}3.5$\,s shift to guard transient accelerations, then relaxes below the static line during steady motion so the quadrotor can commit through transient gaps.

Separating speed classes addresses a different failure mode, as slow and fast obstacles carry asymmetric acceleration risk: near-stationary obstacles can abruptly restart or reverse, whereas cruising obstacles mostly hold velocity.
Pooling them into a single bound therefore breaks in opposite directions across regimes.
Under $\SH$, abundant fast-obstacle updates pull the shared bound downward, leaving rare slow obstacles badly under-covered ($59\,\%$ slow-class misses in Crowd, whereas per-class tracking keeps unobserved slow obstacles guarded at $q_0{=}1$ in Fig.~\ref{fig:q_traj_3d}(a)).
Under $\Sterm$, frequent slow-reversal spikes pull the shared bound upward and over-inflate fast-obstacle capsules, causing \emph{Ours (Pooled Single $q$)} to drop $8.16$\,pp overall and lose nearly half its completion in \emph{Motion Shift}.

Certifying over $[0, H]$ instead of $[0, t^\star]$ (\emph{Ours w/o EW}) forces the filter to hedge against obstacle motion long after the quadrotor has already reached $\Sterm$.
Because capsule radius grows quadratically in lag time, that redundant tail inflates capsules, doubles Tracker Miss, and cuts completion by $27.25$\,pp.
Even with truncation, abrupt non-yielding reversals occasionally exceed $q_{\max}{=}1$ and trigger upper clipping (roughly $19\,\%$ of commits), explaining the $17.02\,\%$ nominal Plan Miss in Proposition~\ref{prop:aci}.
In closed loop, however, the filter re-commits on over $95\,\%$ of control steps (mean execution interval $1.7$ steps vs.\ exposure window $89$ steps), discarding unexecuted tails long before late-lag breaches occur; Execute Miss therefore drops by more than two orders of magnitude to $0.093\,\%$ with zero collisions, confirming the hierarchy~\eqref{eq:hierarchy}.

\subsection{Coupling Between Online Calibration and Learned Backups}\label{sec:exp_coupling}

Prior work typically studies backup learning or occupancy tightening separately.
To test when the two mechanisms interact, Table~\ref{tab:coupling_2x2_3d} crosses calibration state ($\{q{=}1, \text{Online}\}$) against backup policy ($\{\text{Analytic}, \pi_b^{\mathrm{NN}}\}$) inside the same Isaac Sim 3D simulator, measuring their interaction via $\Delta_{\mathrm{int}} = R_{11} - R_{10} - R_{01} + R_{00}$.

\begin{table}[t]
\centering
\caption{Calibration--backup coupling ablation in Isaac Sim 3D (zero collisions throughout).}\label{tab:coupling_2x2_3d}
\scriptsize\setlength{\tabcolsep}{3.2pt}
\resizebox{\columnwidth}{!}{%
\begin{tabular}{@{}llccc@{}}
\toprule
Condition (Isaac Sim 3D) & Backup $\pi_b$ & No Calib.\ ($q{=}1$) & Online Calib. & Calib.\ Gain \\
\midrule
\textbf{Regime~I} ($\SH$, Crowd) & Analytic & 83.1\,\% & 95.0\,\% & $+11.9$\,pp \\
\textbf{Regime~I} ($\SH$, Crowd) & Learned ($\pi_b^{\mathrm{NN}}$) & 83.4\,\% & 95.7\,\% & $+12.3$\,pp \\
\multicolumn{2}{@{}l}{\emph{Interaction $\Delta_{\mathrm{int}}$ under $\SH$}} & \multicolumn{3}{c}{\textbf{$+0.4$\,pp (nearly decoupled)}} \\
\midrule
\textbf{Regime~II} ($\Sterm$, 4 scenes) & Analytic & 0.00\,\% & 38.54\,\% & $+38.54$\,pp \\
\textbf{Regime~II} ($\Sterm$, 4 scenes) & Learned ($\pi_b^{\mathrm{NN}}$) & 0.69\,\% & \textbf{75.69\,\%} & $+75.00$\,pp \\
\multicolumn{2}{@{}l}{\emph{Interaction $\Delta_{\mathrm{int}}$ under $\Sterm$}} & \multicolumn{3}{c}{\textbf{$+36.46$\,pp (strong coupling)}} \\
\bottomrule
\end{tabular}%
}\\[1pt]
{\footnotesize All configurations use the same Isaac Sim 3D quadrotor simulator, LiDAR sensing, mismatch tubes, and exposure-window certifier.}
\vspace{-4pt}
\end{table}

Under $\SH$, calibration and backup choice decouple ($\Delta_{\mathrm{int}} {=} +0.4$\,pp).
Because braking already reaches $\SH$ anywhere clear of walls, tightening obstacle capsules around a fixed stopping trajectory yields the same $\approx +12$\,pp recovery for both analytic and learned backups; once those capsules shrink, straight braking already certifies almost every nominal transition, leaving little extra space for a learned backup to exploit.

Under $\Sterm$, the two mechanisms become strongly interdependent ($\Delta_{\mathrm{int}}{=}+36.46$\,pp vs.\ $+0.4$\,pp under $\SH$).
Shrinking occupancy is necessary to open a traversable corridor window, yet insufficient if the backup cannot maneuver through that window into $\Sterm$.
Without calibration ($q{=}1$), no certified corridor gap exists for either backup.
Turning on calibration with an analytic backup clears straight-through crossings ($38.54\,\%$), but still stalls completely in \emph{Side Pocket} ($0.0\,\%$, Fig.~\ref{fig:isaac_3d_schematic}(b)) because straight braking cannot steer laterally into the recessed bay while the shuttle passes.
Pairing online calibration with $\pi_b^{\mathrm{NN}}$ supplies both the shrunk occupancy window and the lateral refuge-entry maneuvers, lifting completion to $75.69\,\%$ while shortening mean $t^\star$ from $2.194$\,s to $1.512$\,s.

\subsection{Comparison with Baselines Across Conditions and Dynamics}\label{sec:exp_baselines}

\begin{table*}[t]
\centering
\caption{Baseline comparison across terminal-set conditions and simulators.}\label{tab:baselines_2x2}
\scriptsize\setlength{\tabcolsep}{3.6pt}
\resizebox{\textwidth}{!}{%
\begin{tabular}{@{}llccccc@{}}
\toprule
Condition \& Simulator & Method & Overall Comp.\ (\%) & Collisions & Key Metric / Per-Scene Breakdown & Plan Miss & Execute Miss \\
\midrule
\multicolumn{7}{@{}l}{\emph{Regime~I ($\SH$, Yielding) --- Planar 2D Double Integrator (9 scenarios)}} \\
 & NavRL (Unfiltered) & 76.11 $\pm$ 2.00 & \textbf{344} & Feas.\ 85.62\,\% (343 dyn / 1 wall) & N/A & N/A \\
 & NavRL + Safe Shield (ORCA) & 72.22 $\pm$ 1.82 & \textbf{230} & Feas.\ 81.25\,\% (29 dyn / 201 wall) & N/A & N/A \\
 & Gatekeeper-2D (Full-$H$, $q{=}1$) & 11.11 $\pm$ 0.00 & 0 & Feas.\ 12.50\,\%, $t^\star{=}3.50$\,s & 0.00\,\% & --- \\
 & \textbf{Ours} & \textbf{88.89 $\pm$ 0.00} & \textbf{0} & Feas.\ \textbf{100.0\,\%}, $t^\star{=}1.52$\,s & 3.95\,\% & --- \\
\midrule
\multicolumn{7}{@{}l}{\emph{Regime~I ($\SH$, Yielding) --- Isaac Sim 3D Quadrotor (5 scenarios: Crowd / Corr.\ / Open / Purs.\ / Cross)}} \\
 & NavRL (Unfiltered) & 95.5 (Crowd) & 39--411 & Corr.\ 32.0 / Open 88.0 / Purs.\ 82.5 / Cross 91.4 & N/A & N/A \\
 & CBF-QP (Uncertified) & 97.3 (Crowd) & 1--297 & Corr.\ 49.7 / Open 93.0 / Purs.\ 86.5 / Cross --- & N/A & N/A \\
 & Gatekeeper-3D ($q{=}1$) & 83.1 (Crowd) & 0 & Corr.\ 18.1 / Open 62.7 / Purs.\ 48.0 / Cross 77.4 & 0.00\,\% & --- \\
 & \textbf{Ours} & \textbf{95.6} (Crowd) & \textbf{0} & Corr.\ \textbf{20.4} / Open 57.5 / Purs.\ \textbf{52.0} / Cross \textbf{88.6} & \textbf{0.34--0.995\,\%} & --- \\
\midrule
\multicolumn{7}{@{}l}{\emph{Regime~II ($\Sterm$, Non-Yielding) --- Planar 2D Double Integrator (9 scenarios)}} \\
 & NavRL (Unfiltered) & 76.11 $\pm$ 2.00 & \textbf{344} & Feas.\ 85.62\,\% (343 dyn / 1 wall) & N/A & N/A \\
 & NavRL + Safe Shield (ORCA) & 72.22 $\pm$ 1.82 & \textbf{230} & Feas.\ 81.25\,\% (29 dyn / 201 wall) & N/A & N/A \\
 & Gatekeeper-2D ($\abar{=}0.50$) & 78.06 $\pm$ 0.49 & 0 & Feas.\ 87.81\,\%, $t^\star{=}3.32$\,s & 6.70\,\% & --- \\
 & \textbf{Ours} & \textbf{82.22 $\pm$ 1.10} & \textbf{0} & Feas.\ \textbf{92.50\,\%}, $t^\star{=}1.31$\,s & \textbf{3.91\,\%} & --- \\
\midrule
\multicolumn{7}{@{}l}{\emph{Regime~II ($\Sterm$, Non-Yielding) --- Isaac Sim 3D Quadrotor (4 scenarios)}} \\
 & NavRL (Unfiltered) & 76.22 $\pm$ 0.30 & \textbf{137} & SP 26.4 / CR 92.4 / SG 92.4 / MS 93.8 & N/A & N/A \\
 & NavRL + Safe Shield ($r{=}0.60$\,m) & 97.22 $\pm$ 1.59 & \textbf{10} & SP 94.4 / CR 97.9 / SG 98.6 / MS 97.9 & N/A & N/A \\
 & NavRL + Safe Shield ($r{=}0.40$\,m) & 92.01 $\pm$ 2.10 & \textbf{46} & SP 91.0 / CR 93.1 / SG 93.1 / MS 91.0 & N/A & N/A \\
 & Gatekeeper-3D ($\abar{=}0.50$) & 0.00 $\pm$ 0.00 & 0 & SP 0.0 / CR 0.0 / SG 0.0 / MS 0.0 ($t^\star{=}2.19$\,s) & 3.15\,\% & --- \\
 & Gatekeeper-3D ($\abar{=}0.25$) & 54.69 $\pm$ 0.52 & 0 & SP 0.0 / CR 91.0 / SG 87.5 / MS 40.3 ($t^\star{=}2.19$\,s) & 40.12\,\% & 0.541\,\% \\
 & \textbf{Ours} & \textbf{75.69 $\pm$ 1.20} & \textbf{0} & SP \textbf{63.9} / CR \textbf{93.1} / SG \textbf{88.2} / MS \textbf{57.6} ($t^\star{=}1.51$\,s) & \textbf{17.02\,\%} & \textbf{0.093\,\%} \\
\bottomrule
\end{tabular}%
}\\[1pt]
{\footnotesize All certified methods incur zero collisions across every setting. Uncertified baselines (NavRL, CBF-QP, ORCA Safe Shield) incur collisions because they omit bounded-mismatch dynamics or terminal-set invariance.}
\vspace{-2pt}
\end{table*}

Having isolated each mechanism inside Isaac Sim 3D, we next ask how the full filter compares against reactive shields and fixed-budget certified baselines, and whether the observed trends depend on 3D quadrotor lag or carry over from planar 2D dynamics (Table~\ref{tab:baselines_2x2}).

Reactive baselines (unfiltered NavRL, ORCA Safe Shield, CBF-QP) enforce only local single-step avoidance without certifying a bounded-mismatch rollout into an invariant terminal set.
When obstacles reverse or pinch narrow passages, these myopic filters enter states from which no collision-free recovery exists, causing collisions across all four settings in Table~\ref{tab:baselines_2x2}.
Conversely, Gatekeeper-2D/3D relies on full-horizon certification with a fixed acceleration budget: conservative budgets ($q{=}1$ or $\abar{=}0.50$) induce severe deadlocks, whereas manually lowering $\abar$ to $0.25$ in 3D still fails in \emph{Side Pocket} due to its analytic backup and incurs much higher Execute Miss than \emph{Ours} ($0.541\,\%$ vs.\ $0.093\,\%$) because a fixed bound cannot react to sudden spikes.

Across all four settings, \emph{Ours} remains collision-free while outperforming Gatekeeper-2D/3D (Table~\ref{tab:baselines_2x2}).
More importantly, under $\Sterm$ the completion margin over Gatekeeper widens sharply from $+4.17$\,pp in planar 2D to $+21.01$\,pp in Isaac Sim 3D.
This difference reflects dynamics fidelity: under exact 2D kinematics, zero mismatch tubes leave generous geometric slack for a fixed-budget filter, whereas 3D attitude lag introduces non-zero error tubes $(e_p, e_v)$ that consume corridor clearance, so exposure-window truncation, online bound adaptation, and maneuver-capable learned backups become decisive in realistic 3D flight.

\section{Conclusion and Limitations}

Worst-case occupancy inflation over a verification horizon drives backup-filter conservatism, while the terminal-set condition over the exposure window $[0, t^\star]$ determines which bottleneck dominates and whether calibration and backup learning couple.
Within Isaac Sim 3D, controlled ablations confirm that online per-class exposure-window calibration decouples from backup choice under passive yielding ($\SH$, raising both analytic and learned backups from $\approx 83\,\%$ to $95.0\text{--}95.7\,\%$ at $<1\,\%$ Plan Miss) yet couples strongly with our certificate-aligned learned backup $\pi_b^{\mathrm{NN}}$ under non-yielding traffic ($\Sterm$, lifting zero-collision completion to $75.69\,\%$ with $0.093\,\%$ Execute Miss).
Extending evaluation across both terminal-set conditions and both 2D and 3D simulators confirms consistent zero-collision gains over Gatekeeper-2D/3D without the collisions of reactive safety shields.
Tracking assumes persistent obstacle IDs within $R_s$ and bounded sensing noise $(\varepsilon,\varepsilon_v)$ and certified 3D mismatch tubes hold on $>99.90\,\%$ of steps; whenever obstacle accelerations exceed $q_{\max}{=}1$, upper clipping ($U/T$) relaxes conditional coverage, so zero observed collisions under high-rate replanning is a statistical outcome rather than an unconditional guarantee.
Bounding sensed-feedback bias, confirming coupled exposure-window shaping, and physical quadrotor flight deployment remain future work.

\bibliographystyle{IEEEtran}
\bibliography{refs}

\end{document}